%% file: main.tex
\documentclass[letterpaper]{article} 
\usepackage{aaai2026}  
\usepackage{times}  
\usepackage{helvet}  
\usepackage{courier}  
\usepackage[hyphens]{url}  
\usepackage{graphicx} 
\usepackage{natbib}  
\usepackage{caption} 
\usepackage{algorithm}
\usepackage{algorithmic}

\usepackage{amsfonts}
\usepackage{amsmath}
\usepackage{float}
\usepackage{cuted}

\usepackage{newfloat}
\usepackage{listings}
\DeclareCaptionStyle{ruled}{labelfont=normalfont,labelsep=colon,strut=off} 
\floatstyle{ruled}
\newfloat{listing}{tb}{lst}{}
\floatname{listing}{Listing}
\title{PRevivor: Reviving Ancient Chinese Paintings using Prior-Guided Color Transformers}
\author{
    Tan Tang\textsuperscript{\rm 1},
    Yanhong Wu\textsuperscript{\rm 2},
    Junming Gao\textsuperscript{\rm 1},
    Yingcai Wu\textsuperscript{\rm 2}\thanks{Corresponding author.}
}
\affiliations{
    \textsuperscript{\rm 1}Laboratory of Art and Archaeology Image, Zhejiang University, Tianmushan Road No. 148, Hangzhou, China\\
    \textsuperscript{\rm 2}State Key Lab of CAD\& CG, Zhejiang University, Yuhangtang Road No. 866, Hangzhou, China

}

\usepackage{bibentry}

\begin{document}

\maketitle

\begin{abstract}
Ancient Chinese paintings are a valuable cultural heritage that is damaged by irreversible color degradation.
Reviving color-degraded paintings is extraordinarily difficult due to the complex chemistry mechanism.
Progress is further slowed by the lack of comprehensive, high-quality datasets, which hampers the creation of end-to-end digital restoration tools.
To revive colors, we propose PRevivor, a prior-guided color transformer that learns from recent paintings (e.g., Ming and Qing Dynasty) to restore ancient ones (e.g., Tang and Song Dynasty).
To develop PRevivor, we decompose color restoration into two sequential sub-tasks: luminance enhancement and hue correction. 
For luminance enhancement, we employ two variational U-Nets and a multi-scale mapping module to translate faded luminance into restored counterparts.
For hue correction, we design a dual-branch color query module guided by localized hue priors extracted from faded paintings.
Specifically, one branch focuses attention on regions guided by masked priors, enforcing localized hue correction, whereas the other branch remains unconstrained to maintain a global reasoning capability.
To evaluate PRevivor, we conduct extensive experiments against state-of-the-art colorization methods.
The results demonstrate superior performance both quantitatively and qualitatively.
\end{abstract}


\input{1_introduction}
\input{2_relatedwork}
\input{3_framework}
\input{4_experiment}
\input{5_conclusion}

\bibliography{aaai2026}

\end{document}

%% file: 1_introduction.tex
\section{Introduction}
Color degradation steadily damages cultural heritage of its original brilliance. 
Over centuries, pigments in ancient paintings darken as oxidation alters their chemistry.
Although chemists and conservators have devised sophisticated treatments to slow this decline~\cite{Ruiz2021a, Haslam2020a}, no intervention can fully reverse the loss; the vivid hues laid out first by the artist remain forever out of reach.
High resolution digitization offers a powerful alternative to physical restoration: advanced image processing algorithms~\cite{BWei2003a,Chen2013a} could change the damaged colors of the painting in the digital realm without risking the fragile original.
However, these conventional approaches require extensive expert knowledge and do not yet fully revive the original colors.

Recent advances in deep learning have largely resolved this limitation: by extracting color-restoration patterns from vast datasets.
Successful applications could be found in restoring old photos~\cite{wan2020bringing, jin2021focusing} or recoloring black images~\cite{cheng2015deep, xia2022disentangled, kumar2021colorization, ji2022colorformer, weng2022ct, kang2023ddcolor, kim2022bigcolor}.
However, heritage artifacts pose unique problems: localized abrasions, overlapping stains, and chemically altered pigments create degradation patterns that current automatic colorization models fail to learn.
Interactive colorization resolve parts of the limits, but requiring extra expert guidance, which is too slow and labor-intensive for large collections.

To bridge the gap, we introduce PRevivor, an innovative image colorization framework specifically designed to restore ancient Chinese silk paintings.
However, creating such a framework faces two significant hurdles.
First, the scarcity of high-quality paired datasets of ancient and modern paintings hampers the direct learning of color degradation patterns through conventional data-driven methods. 
Existing benchmark datasets, sourced from diverse and unrelated domains, often fail to capture the unique vibrancy and nuances required for the accurate restoration of ancient paintings. 
Second, the intricate and implicit degradation patterns of ancient paintings are difficult to model accurately. 
It remains unclear which neural network architecture is best suited for effectively reviving the original colors and details of these historical artworks.

To address the first challenge, we decompose the color restoration of paintings into luminance enhancement and hue correction. 
Based on domain-specific observations, we design two empirically driven degradation simulation methods.
We selected well-preserved paintings (e.g., thoese from Ming and Qing dynasties) and degrade them to imitate older ones (e.g., thoese from Tang and Sont dynasties).
In addition, a domain translation approach~\cite{wan2020bringing} is employed to ensure the alignment of the synthetic and real data distributions in the latent space.
To address the second challenge, we adapt an automatic image colorization approach~\cite{kang2023ddcolor} to accept residual color priors present in the degraded painting. 
Furthermore, we introduce a mask-guided pixel loss that constrains the hue correction process based on valid color priors.

To evaluate PRevivor, we curate a specialized dataset of over 300 degraded and 870 well-preserved ancient Chinese painting patches from museums and digital archives.
We conduct comprehensive comparisons against six state-of-the-art colorization methods across both paired and unpaired evaluation settings. 
Our assessment employs quantitative metrics (PSNR, SSIM, LPIPS, FID, and $\Delta$Colorfulness) and qualitative visual analysis, supplemented by human evaluation with 8 expert art restoration specialists across 35 test paintings.
Qualitative results demonstrate PRevivor's advantages in preserving authentic color distributions and maintaining cultural authenticity through balanced luminance recovery.
The results show the superior performance of PRevivor in technical and perceptual measures, achieving ~$55\%$ median expert preference and optimal quantitative metrics, particularly excelling in color accuracy ($\Delta$Colorfulness: 1.70) and distributional alignment (FID: 61.95) essential for authentic restoration of ancient Chinese paintings.

%% file: 2_relatedwork.tex
\section{Related Works}
In this section, we review existing approaches across three key areas: automatic image colorization methods, cultural heritage and historical image restoration techniques, and interactive colorization frameworks. While these approaches have achieved significant progress, they lack mechanisms for incorporating historical context and cultural authenticity required for ancient artwork colorization.

\subsection{Automatic Image Colorization}
Automatic image colorization aims to add plausible colors to grayscale images without additional user guidance. Early deep learning approaches faced challenges with color ambiguity and under-saturation issues.

\textbf{CNN-based Methods}:
Cheng et al.~\cite{cheng2015deep} proposed the first deep neural network approach but suffered from over-smoothing due to regression losses. Zhang et al.~\cite{zhang2016colorful} addressed this by formulating colorization as a classification problem in quantized CIELAB color space with cross-entropy loss. InstColor~\cite{su2020instance} incorporated object detection to reduce color overflow by colorizing individual instances separately. DISCO~\cite{xia2022disentangled} introduced a coarse-to-fine framework using anchor points to learn global color affinity. However, these CNN-based methods rely on modern datasets with contemporary color patterns, failing to capture historical pigment limitations and cultural color symbolism essential for ancient paintings.

\textbf{Transformer-based Methods}:
ColTran~\cite{kumar2021colorization} presented the first transformer-based colorization model with multi-stage processing. ColorFormer~\cite{ji2022colorformer} introduced Global-Local hybrid self-attention and color memory modules. CT2~\cite{weng2022ct} leveraged meaningful color tokens based on color space statistics. DDColor~\cite{kang2023ddcolor} employed learnable color tokens with cross-attention mechanisms. Despite advanced architectures, these methods lack mechanisms for incorporating art historical knowledge and cannot handle colors in historical contexts.

\textbf{GAN-based Methods}: 
CycleGAN~\cite{zhu2017unpaired} by Zhu et al. introduced cycle-consistent adversarial training for unpaired domain translation.
ChromaGAN~\cite{vitoria2020chromagan} utilized PatchGAN discriminators with WGAN loss.
Wan et al.~\cite{wan2020bringing} proposed a triplet domain translation network specifically for severely degraded historical photos, addressing the domain gap between synthetic training data and real degraded images.
HistoryNet~\cite{jin2021focusing} specifically targeted historical images with classification and segmentation information. BigColor~\cite{kim2022bigcolor} leveraged pre-trained BigGAN generators but required substantial computational resources. 
While promising for style transfer, GAN-based methods may generate visually plausible but historically inaccurate colors, lacking explicit mechanisms for cultural authenticity and expert art historical guidance.

\textbf{Diffusion-based Methods}: 
Recent diffusion models like Palette~\cite{saharia2022palette} and ColorDiff~\cite{wang2024multimodal} have demonstrated superior generative capabilities but suffer from long inference times due to iterative denoising processes. The stochastic nature of diffusion processes lacks the precision required for historical accuracy, and current models are trained on modern image distributions that do not reflect historical artistic practices.

\subsection{Cultural Heritage and Historical Image Restoration}
The restoration of cultural heritage and historical imagery presents unique challenges requiring specialized approaches beyond standard image restoration techniques. Traditional approaches focused primarily on damage detection and inpainting-based repair, but these methods were limited to addressing localized defects and could not handle the complex, mixed degradations common in historical artifacts.

Recent deep learning approaches have addressed various aspects of historical image restoration. 
Wang et al.~\cite{wang2018dunhuang} created a systematic restoration framework for high-resolution deteriorated mural textures, addressing GPU memory limitations through patch-based processing combined with GP-Editing techniques to ensure color consistency. More recently, the MER model~\cite{wei2025progressive} introduced a two-stage approach combining illumination enhancement with automatic defect detection using multi-receptive field strategies, specifically targeting the challenges of low-light conditions during image capture at archaeological sites.
However, these restoration-focused methods primarily address degradation repair rather than color reconstruction, lacking mechanisms for historically accurate color assignment and integration with art historical knowledge essential for reviving ancient paintings.

\subsection{Interactive Colorization}
Interactive colorization methods enable precise user control through cross-modal models that establish correspondence between different modalities. Early approaches built cross-modal correspondence from scratch using manually annotated datasets~\cite{weng2022code}, transformer architectures with pre-trained language models~\cite{chang2022coder}, and grouping mechanisms~\cite{chang2023coins}, but required extensive paired text-image datasets unavailable for historical artworks. Recent methods leverage pre-trained cross-modal models like CLIP~\cite{huang2022unicolor} and Stable Diffusion through various conditioning paradigms including mid-layer feature insertion~\cite{weng2023cad}, ControlNet-based encoders~\cite{li2024coco,liang2024control}, and direct grayscale input~\cite{zabari2023diffusing}. However, these pre-trained models are trained on contemporary data that may not reflect historical artistic vocabulary and can produce culturally inappropriate color combinations violating historical artistic principles.

%% file: 3_framework.tex
\section{PRevivor Framework}
Given a color-degraded painting image $\mathbf{x}_{\mathrm{Lab}} \in \mathbb{R}^{H \times W \times 3}$, our objective is to recover a color-restored painting image $\mathbf{y}_{\mathrm{Lab}} \in \mathbb{R}^{H \times W \times 3}$.
Due to the lack of high-quality paired data, end-to-end supervised learning cannot be applied directly.
To alleviate this problem, we decomposes the restoration process into two sequential stages, specifically luminance enhancement and hue correction.

\begin{figure*}[htbp]
    \centering
    \includegraphics[width=0.9\textwidth,height=0.33\textheight,keepaspectratio]{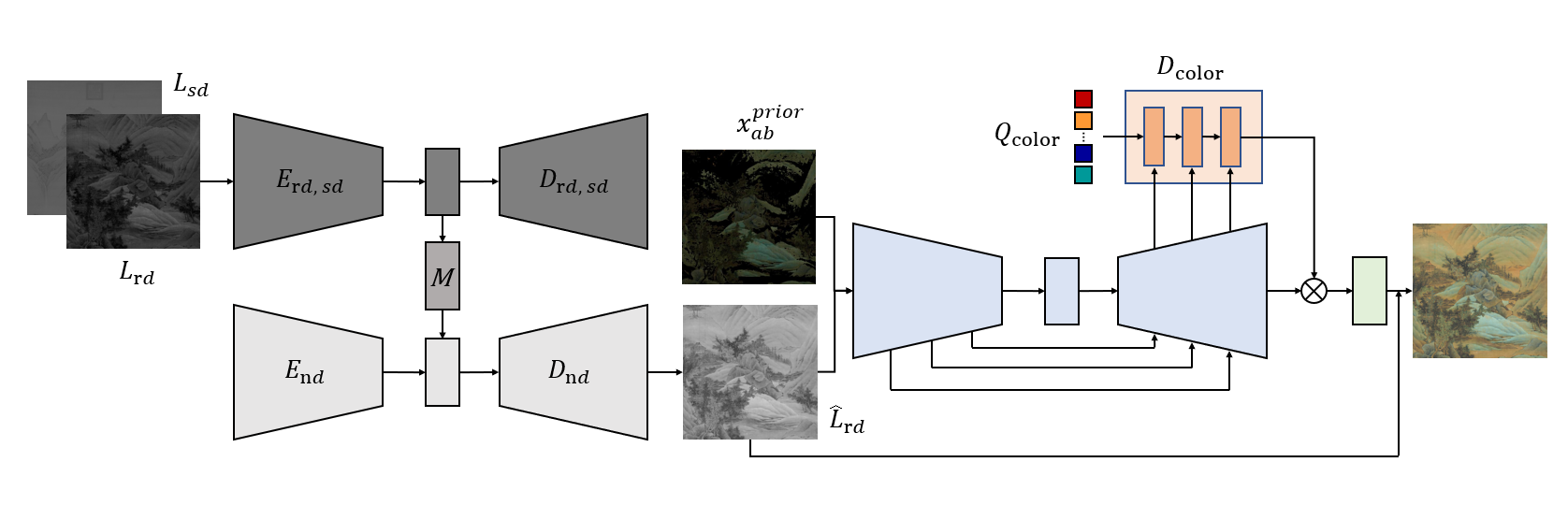}
    \caption{Overview of the proposed PRevivor framework. The framework consists of two sequential stages: luminance enhancement and hue correction. The luminance enhancement stage employs domain translation to bridge the gap between real and synthetic degraded data, while the hue correction stage leverages extracted color priors for accurate restoration.}
    \label{fig:framework}
\end{figure*}

At first, we adopt a hybrid strategy that combines real-world luminance degradation curves with empirically designed linear degradation curves to simulate the luminance degradation.
Moreover, we employ a domain translation approach~\cite{wan2020bringing} to bridge the gap between real and synthetic data.
Secondly, we propose a color-prior extraction method and modify an advanced automatic image colorization approach~\cite{kang2023ddcolor}, enabling it to leverage the extracted prior for more accurate hue correction.

\subsection{Luminance Enhancement}
The luminance enhancement involves image translation across three domains: real degraded luminance images $\mathbf{L}_{\mathrm{rd}}$, synthetic degraded luminance images $\mathbf{L}_{\mathrm{sd}}$, and real non-degraded luminance images $\mathbf{L}_{\mathrm{nd}}$.
To generate synthetic degraded luminance images, we first extracted real-world luminance degradation curves from a small collection of paired painting images that had undergone restoration. 
Specifically, we computed bin-wise statistics of per-pixel luminance differences between the degraded and restored paintings.
By observing the extracted real-world luminance degradation curves, we were surprised to find that they are overall close to linear. Therefore, we further designed linear degradation curves to simulate a wider range of luminance degradation:
\begin{equation}
\mathbf{L}_{\mathrm{sd}} = \alpha \times \mathbf{L}_{\mathrm{nd}} + \beta
\end{equation}
where \(\alpha\) is randomly sampled from the interval \([0.2, 0.5]\), and \(\beta\) is randomly sampled from \([15, 25]\).

The domain translation approach consists of three modules: two variational autoencoders and a mapping network~\cite{wan2020bringing}.
The first variational autoencoder $\{\mathbf{E}_{rd, sd}, \mathbf{D}_{rd, sd}\}$ learns a latent space shared between $\mathbf{L}_{\mathrm{rd}}$ and $\mathbf{L}_{\mathrm{sd}}$, whereas the second one $\{\mathbf{E}_{nd}, \mathbf{D}_{nd}\}$ models the latent space of $\mathbf{L}_{\mathrm{nd}}$ independently.
The mapping network $\mathbf{M}$ is used to transform the latent space learned by the first variational autoencoder into the latent space of the second variational autoencoder.
Given a real degraded luminance image $\mathbf{L}_{\mathrm{rd}}$, its restored luminance image $\hat{\mathbf{L}}_{\mathrm{rd}}$ can be obtained as follows:
\begin{equation}
\hat{\mathbf{L}}_{\mathrm{rd}} = \mathbf{D}_{\mathrm{nd}}\left( \mathbf{M}\left( \mathbf{E}_{\mathrm{rd, sd}}\left( \mathbf{L}_{\mathrm{rd}} \right) \right) \right)
\end{equation}

To train the first variational autoencoder, samples are randomly drawn from $\mathbf{L}_{\mathrm{rd}}$ and $\mathbf{L}_{\mathrm{sd}}$ with equal probability.
Additionally, the two types of degradation curves are selected with equal probability to generate $\mathbf{L}_{\mathrm{sd}}$ during training.
The final losses for training the first variational autoencoder are adopted as:
\begin{equation}
\begin{split}
\mathcal{L}_{\mathbf{E}_{rd, sd}, \mathbf{D}_{rd, sd}} =\ 
& \mathcal{L}_{\text{pixel}} + \mathcal{L}_{\text{GAN}} 
+ \mathcal{L}_{\text{GAN-feat}} \\
& + \mathcal{L}_{\text{GAN-latent}} + \mathcal{L}_{\text{KL-latent}}
\end{split}
\end{equation}

where $\mathcal{L}_{\text{pixel}}$ is the pixel-level reconstruction loss, implemented using Smooth L1 Loss to minimize the difference between the input and generated images.
$\mathcal{L}_{\text{GAN}}$ is the adversarial loss that encourages the autoencoder to produce images indistinguishable from the input ones.
$\mathcal{L}_{\text{GAN-feat}}$ denotes the feature-level adversarial loss, which compares intermediate features from the discriminator to ensure structural and textural consistency.
$\mathcal{L}_{\text{GAN-latent}}$ is the adversarial loss in the latent space, designed to encourage similarity between $\mathbf{L}_{\mathrm{rd}}$ and $\mathbf{L}_{\mathrm{sd}}$.
$\mathcal{L}_{\text{KL-latent}}$ denotes the KL divergence loss, which regularizes the latent distribution to match a standard Gaussian prior.
The only loss difference in the second variational autoencoder is the absence of the $\mathcal{L}_{\text{GAN-latent}}$, since it operates solely on $\mathbf{L}_{\mathrm{nd}}$.

Compared to the two autoencoders, the loss of the mapping network is defined as:
\begin{equation}
\begin{split}
\mathcal{L}_{M} =\ 
& \mathcal{L}_{\text{pixel}} + \mathcal{L}_{\text{GAN}} 
+ \mathcal{L}_{\text{GAN-feat}} + \mathcal{L}_{\text{latent}},
\end{split}
\end{equation}

where $\mathcal{L}_{\text{latent}}$ is constructed as
\begin{equation}
\mathcal{L}_{\text{latent}} = \left\| M\left( \mathbf{E}_{\mathrm{rd, sd}}\left( \mathbf{L}_{\mathrm{rd}} \right) \right) - \mathbf{E}_{\mathrm{nd}}\left( \mathbf{L}_{\mathrm{nd}} \right) \right\|_1
\end{equation}

\subsection{Hue Correction}
Once the luminance of the painting image is fixed through enhancement, the subsequent step in color restoration involves correcting the degraded hue.
Specifically, we adopt a pretrained ConvNeXt~\cite{liu2022convnet} as the encoder, followed by a U-Net decoder ($\mathbf{D}_{U-Net}$) to extract multiscale image features.
Compared with DDColor~\cite{kang2023ddcolor}, our encoder takes the full Lab image as input, rather than only the L channel, where the additional ab channels serve as color priors extracted from the color-degraded painting image.
The color prior extraction process is comprised of four steps. 
First, we applied a background removal model to extract the painting’s background, i.e., silk used in Chinese paintings.
Considering that the model may mistakenly remove part of the foreground in some cases, we further introduced a threshold-based pixel filtering mechanism based on prior knowledge of silk colors.
We then assumed that the silk region is both gradient-smooth and dominant in the background, and employed image gradients combined with K-means clustering to estimate the representative silk color $\mathbf{c}_{\mathrm{silk}}$.
Finally, we computed the per-pixel ab-channel distance between the entire image and $\mathbf{c}_{\mathrm{silk}}$, retaining only those pixels whose distances exceed a predefined threshold $\tau$ as valid color priors:
\begin{equation}
\mathbf{x}_{\mathrm{ab}}^{\mathrm{prior}} = \mathbf{M}_{\mathrm{prior}} \odot \mathbf{x}_{\mathrm{ab}}
\end{equation}

where $\mathbf{M}_{\mathrm{prior}}$ is defined as
\begin{equation}
\mathbf{M}_{\mathrm{prior}}(i,j) = 
\begin{cases}
1, & \text{if } \left\| \mathbf{x}_{\mathrm{ab}}(i,j) - \mathbf{c}_{\mathrm{silk}} \right\|_2 > \tau \\
0, & \text{otherwise}
\end{cases}
\end{equation}

Since real non-degraded luminance images $\mathbf{L}_{\mathrm{nd}}$ do not contain degraded color priors, we apply a simple linear attenuation to simulate the degradation of the ab channels:
\begin{equation}
\tilde{\mathbf{x}}_{\mathrm{ab}}^{\mathrm{prior}} = \gamma \, \mathbf{x}_{\mathrm{ab}}^{\mathrm{prior}}
\end{equation}

where $\gamma$ is randomly sampled from $[0.2, 0.5]$ for negative $\mathbf{x}_{\mathrm{ab}}^{\mathrm{prior}}$ values, and from $[0.5, 0.9]$ for positive ones.

The extracted multiscale image features are then fed into a multiscale color decoder ($\mathbf{D}_{color}$) along with a series of color queries ($\mathbf{Q}_{color}$). Each decoder block consists of a cross-attention layer, a self-attention layer, and an MLP, arranged in an order empirically determined through performance comparison. The training loss of the entire hue correction module is defined as:
\begin{equation}
\begin{split}
\mathcal{L}_{hue} =\ 
& \mathcal{L}_{\text{pixel}} + \mathcal{L}_{\text{GAN}} + \mathcal{L}_{\text{perceptual}} \\
& + \mathcal{L}_{\text{colorful}} + \mathcal{L}_{\text{mask}}
\end{split}
\end{equation}

where $\mathcal{L}_{\text{perceptual}}$ is the perceptual loss, computed by comparing high-level features extracted from a pretrained VGG network, encouraging the generated images to be perceptually similar to the ground truth.
$\mathcal{L}_{\text{colorful}}$ is the colorful loss, designed to encourage vibrant and diverse color distributions in the generated images by penalizing overly desaturated or dull outputs.
$\mathcal{L}_{\text{mask}}$ is the pixel-mask loss, which applies pixel-wise reconstruction loss only within masked regions, guiding the model to focus its correction efforts on areas where hue priors have been injected.

%% file: 4_experiment.tex
\begin{figure*}[!t]
\centering
\includegraphics[width=0.99\textwidth,height=0.9\textheight,keepaspectratio]{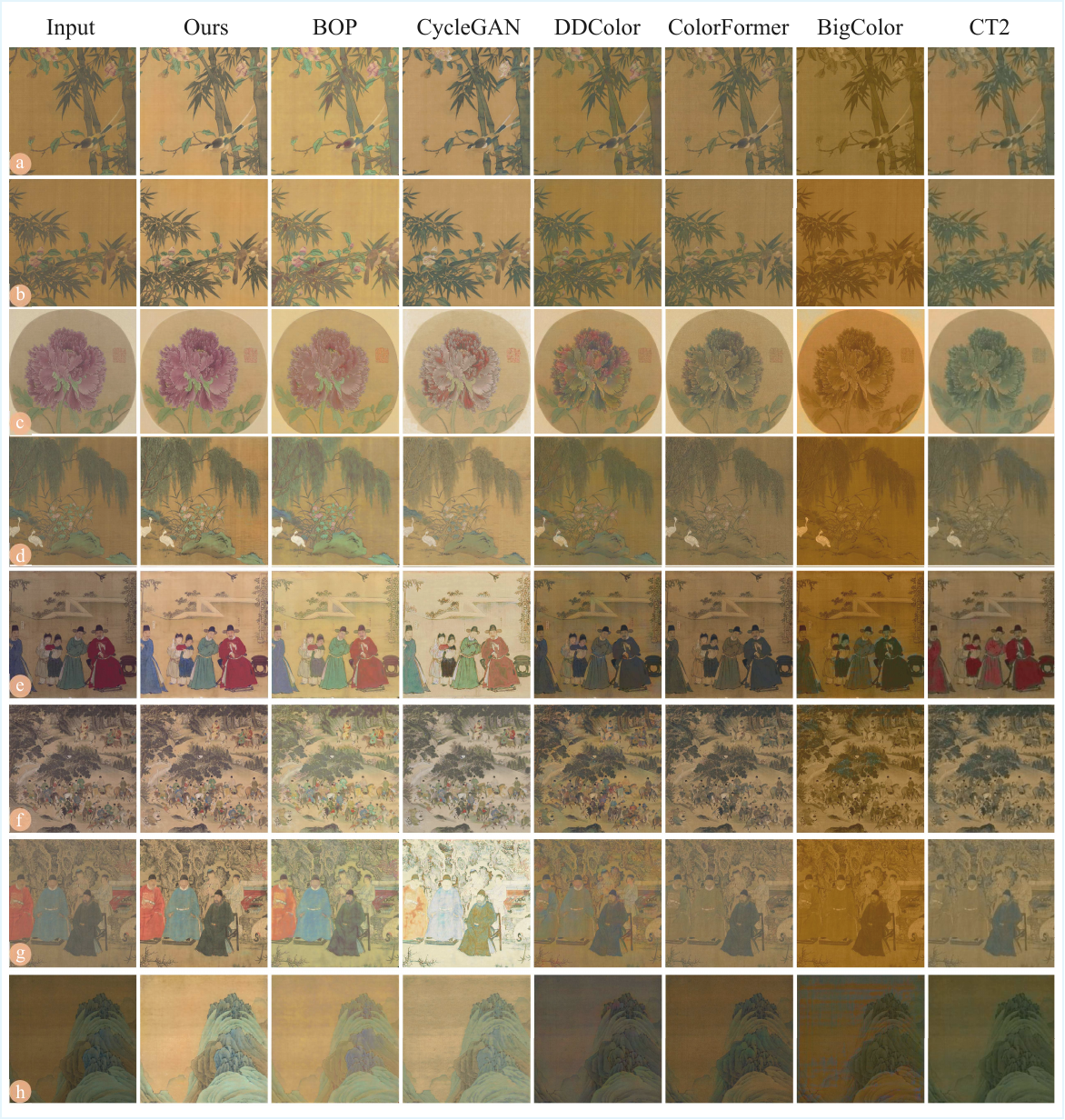}
\caption{Qualitative comparison of color restoration results on diverse ancient Chinese paintings. Our method demonstrates superior color fidelity and luminance enhancement while preserving cultural authenticity across various painting themes including flora, portraits, and landscapes. Zoom in for best view.}
\label{fig:visual_comparison}
\end{figure*}

\section{Experiment}
\subsection{Implementation}
We train our colorization network in a multi-stage manner. Our PRevivor framework consists of a two-stage pipeline trained sequentially. The first stage learns domain translation between degraded and non-degraded luminance data, while the second stage retrains a modified DDColor architecture, incorporating extracted color priors and accepting full Lab input instead of L-channel only.

For the luminance enhancement stage, we sequentially train three components: domain A VAE for degraded images, domain B VAE for restored images, and the mapping network. All VAE models are trained with batch size 32 at 512×512 resolution using Smooth L1 loss without VGG loss. The mapping network employs 6 mapping blocks with feature dimension 512 and L1 feature loss weight of 60.
For the hue correction stage, we use AdamW optimizer with $\beta_1 = 0.9$, $\beta_2 = 0.99$, weight decay = 0.01. The learning rate is initialized to $1 \times 10^{-4}$ and decayed by 0.5 at iterations 4,000, 8,000, 12,000, 16,000, and 20,000. For the loss terms, we set $\lambda_{\text{pix}} = 0.1$, $\lambda_{\text{mask}} = 1.0$, $\lambda_{\text{per}} = 5.0$, $\lambda_{\text{adv}} = 1.0$, and $\lambda_{\text{col}} = 0.5$. We use ConvNeXt-L as the encoder and train for 24,000 iterations with batch size 4 at $512 \times 512$ resolution.

Due to the absence of existing Chinese painting datasets and the challenges posed by long scroll paintings, we curate a specialized dataset for ancient painting restoration. We collect both color-degraded and well-preserved Chinese paintings from museums and digital archives. To address the limited availability of training data while preserving complete visual information, we apply patch-based data augmentation by carefully cropping the collected paintings into smaller segments. This process yields over 300 color-degraded painting patches and over 870 well-preserved painting patches for model training. The degraded samples exhibit various forms of color deterioration including fading, discoloration, and saturation loss, while the preserved samples maintain their original vibrant colors and serve as restoration targets.
We evaluate our method on a curated test set of color-degraded ancient paintings spanning diverse themes and degradation severities. All experiments are conducted on a single NVIDIA A100 GPU with 80GB memory.

\subsection{Comparisons}
We evaluate our approach against 6 representative automatic colorization techniques spanning different paradigms:
(1) Bringing Old Photos Back to Life~\cite{wan2020bringing}; 
(2) CycleGAN~\cite{zhu2017unpaired};
(3) DDColor~\cite{kang2023ddcolor};
(4) BigColor~\cite{kim2022bigcolor};
(5) CT2~\cite{weng2022ct};
(6) ColorFormer~\cite{ji2022colorformer}.
To ensure fair comparison, all baseline methods are retrained on our curated ancient painting dataset using their original training protocols.

\subsubsection{Qualitative Analysis}
Figure~\ref{fig:visual_comparison} presents comprehensive visual evaluations across diverse ancient painting categories. Our framework exhibits notable advantages in both colorization accuracy and luminance enhancement. 
Our method demonstrates superior color fidelity by preserving original hues while restoring vibrancy, as evidenced by the natural reproduction of silk backgrounds and accurate enhancement of traditional pigments. The two-stage pipeline ensures balanced luminance recovery without over-saturation, maintaining the authentic aesthetic characteristics of historical Chinese paintings.
In contrast, competing methods exhibit systematic limitations. BOP (Bring old photos back to life) produces overly vivid results that appear distorted and unnatural. CycleGAN generates excessively bright outputs, causing washed-out appearances (sample g) and color distortions, such as altered red flowers (sample c). DDColor frequently predicts non-existent colors, converting red clothing to black (sample e) or blue (sample g), while its inability to enhance luminance results in persistently dim outputs. ColorFormer, BigColor, and CT2 struggle with effective color restoration and brightness enhancement, likely due to their dependence on large-scale paired datasets (e.g., 1.3M ImageNet pairs~\cite{deng2009imagenet}), which are unavailable for ancient painting restoration tasks.

\textbf{Quantitative Assessment.} 
To establish ground truth for quantitative evaluation, we collaborate with art restoration experts to manually restore 10 severely degraded ancient paintings. We evaluate all methods using four complementary metrics: PSNR and SSIM for pixel-level and structural fidelity, LPIPS for perceptual similarity, and $\Delta$Colorfulness~\cite{hasler2003measuring} to measure color restoration accuracy. We exclude FID as it primarily captures distributional differences rather than restoration quality for individual paintings, making it less suitable for evaluating precise color recovery against expert-restored references.
As shown in Table~\ref{tab:quantitative_results}, our method achieves superior performance across most metrics. We obtain the best PSNR (17.39) and SSIM (0.50), indicating accurate pixel-level reconstruction and structural preservation. Most notably, our approach achieves the lowest $\Delta$Colorfulness (1.70), demonstrating significantly better color restoration compared to the second-best method (BigColor at 3.92). While our LPIPS (0.39) is marginally higher than the best performer, it remains competitive, suggesting minor perceptual differences that are offset by substantial improvements in color accuracy and structural fidelity.

\begin{table*}[!htbp]
\centering
\caption{Quantitative comparison with state-of-the-art colorization methods against expert-restored references. Best results are highlighted in \textbf{bold}.}
\label{tab:quantitative_results}
\begin{tabular}{|l|c|c|c|c|c|}
\hline
Method & $\Delta$Colorfulness $\downarrow$ & LPIPS $\downarrow$ & SSIM $\uparrow$ & PSNR $\uparrow$ \\
\hline
Old Photos Back to Life & 4.05 & \textbf{0.38} & 0.49 & 15.36 \\
CycleGAN & 5.63 & 0.38 & 0.49 & 16.17 \\
BigColor & 3.92 & 0.64 & 0.36 & 10.49 \\
DDColor & 3.92 & 0.46 & 0.45 & 12.68 \\
CT2 & 12.92 & 0.56 & 0.40 & 12.69 \\
ColorFormer & 5.60 & 0.65 & 0.45 & 12.85 \\
\hline
\textbf{Ours} & \textbf{1.70} & 0.39 & \textbf{0.50} & \textbf{17.39} \\
\hline
\end{tabular}
\end{table*}

\begin{table}[!htbp]
\centering
\caption{Quantitative comparison with state-of-the-art colorization methods against well-preserved paintings. Best results are highlighted in \textbf{bold}.}
\label{tab:quantitative_results_unpaired_mask}
\begin{tabular}{|l|c|c|}
\hline
Method & FID $\downarrow$ & $\Delta$Colorfulness $\downarrow$ \\
\hline
Old Photos Back to Life & 112.50 & \textbf{1.88}\\
CycleGAN & 101.31 & 6.66\\
BigColor & 84.56 & 17.69\\
DDColor & 73.56 & 8.52\\
CT2 & 80.67 & 12.66\\
ColorFormer & 84.07 & 14.79\\
\hline
\textbf{Ours} & \textbf{61.95} & 5.82\\
\hline
\end{tabular}
\end{table}

\textbf{Unpaired Evaluation.}
We further evaluate methods against unpaired well-preserved paintings to assess distributional quality and color enhancement capability. Here, we employ FID to measure distributional similarity between restored and well-preserved paintings, alongside $\Delta$Colorfulness to quantify color restoration effectiveness without pixel-level correspondence. To ensure fair comparison, we apply our color prior extraction masks to all methods before metric computation, focusing evaluation on regions where color restoration is most critical and filtering out background silk areas that may confound distributional analysis.
Table~\ref{tab:quantitative_results_unpaired_mask} reveals that our method achieves the best FID (61.95), substantially outperforming competitors and demonstrating superior distributional alignment with authentic well-preserved paintings. This indicates that our restored paintings exhibit color distributions most similar to genuine historical artworks. While our $\Delta$Colorfulness (5.82) is higher than Old Photos Back to Life (1.88), this reflects our method's ability to restore more vibrant colors, which may appear as over-enhancement when compared to potentially faded "well-preserved" references. The FID superiority confirms that our color enhancement aligns better with the authentic color characteristics of historical Chinese paintings.

\textbf{Human Evaluation.} We conduct extensive human assessment using 35 randomly selected paintings from our test collection. Each evaluation session presents participants with all 7 colorization results, displayed in randomized order to eliminate bias. Our participant pool comprises 8 experts in ancient paintings restoration. Evaluators rank results based on three criteria: cultural authenticity, aesthetic appeal, and technical quality.
\begin{figure}[t]
\centering
\includegraphics[width=0.99\columnwidth]{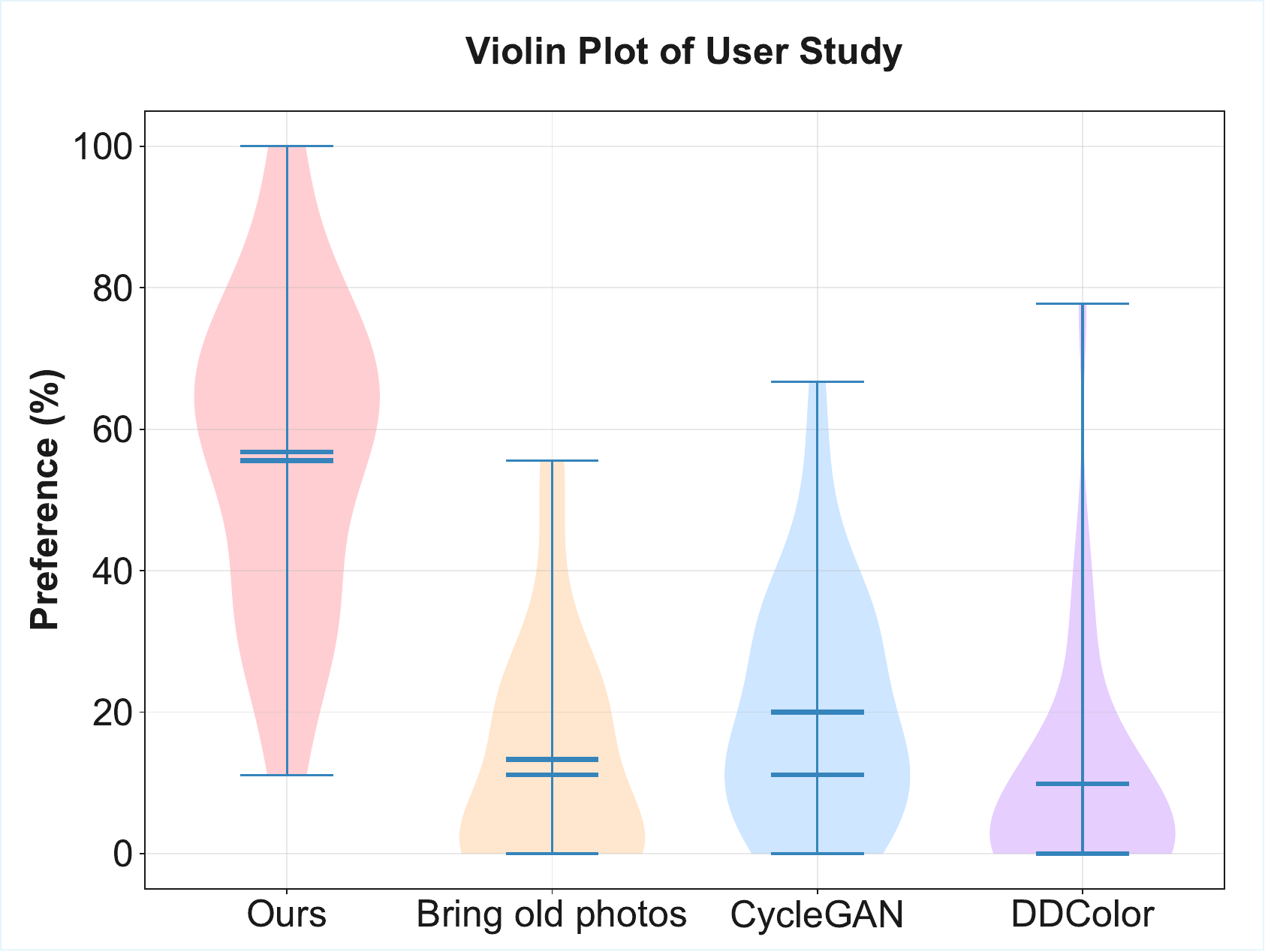}
\caption{Expert preference distribution for top-performing colorization methods. Our method achieves the highest preference scores with strong expert consensus.}
\label{fig:violin_plot}
\end{figure}

Figure~\ref{fig:violin_plot} presents the expert preference distribution for top-performing methods. Our approach achieves remarkable consensus with preference scores concentrated between 50-100\% (median ~55\%), demonstrating consistent high-quality performance across diverse painting styles. In contrast, competing methods show significant limitations: Bring Old Photos exhibits consistently poor performance (0-15\%), CycleGAN displays high variance indicating inconsistent quality (0-65\%), and DDColor shows limited effectiveness (narrow distribution around 10\%) despite its strength in natural image colorization. The substantial preference gap (~40\% median difference) validates our domain-specific design, confirming that the two-stage pipeline effectively addresses ancient Chinese painting restoration challenges.

%% file: 5_conclusion.tex
\section{Conclusion}
While PRevisor demonstrates significant advances in ancient painting restoration, certain limitations warrant future exploration and investigation. Like most automatic colorization methods, our framework lacks fine-grained user control mechanisms for adjusting specific color regions or individual artistic preferences during restoration. Additionally, the two-stage training pipeline, while effective, introduces computational overhead compared to end-to-end approaches. Future work could explore interactive restoration interfaces that allow domain experts to guide the colorization process through sparse annotations, and investigate unified architectures that maintain our method's restoration quality while improving overall training efficiency.

In this work, we propose a prior-guided color transformer, namely PRevivor, to revive color-degraded ancient Chinese paintings.
To overcome data limits, we collected recent paintings from Ming and Qing dynasties and employ a domain translation method to degraded them to imitate older ones.
We then propose a two-stage colorization framework that first enhance luminance and then correct hue to revive degraded colors.
Moreover, we propose a prior calculation method to ensure our model could effectively restore local degradation patterns according to the reserved colors in the paintings.
Finally, we compared PRevivor with recent SOTA models and the results demonstrate that our model outperforms the baseline both quantatively and qualitatively.

    
    
    
    
    
    
    
    
    